\documentclass[twoside,11pt]{article}
\usepackage{jair, hyperref, theapa, rawfonts}

\firstpageno{1}

\usepackage{graphicx}
\usepackage{multirow}
\usepackage{array}
\usepackage{url}

\begin{document}

\title{Recent Trends in Unsupervised Summarization}

\author{\name Mohammad Khosravani \email mkhosrav@lakeheadu.ca \\
        \addr Lakehead University, Ontario, Canada\\
        \AND
        \name Amine Trabelsi \email amine.Trabelsi@usherbrooke.ca \\
        \addr Université de Sherbrooke Quebec, Canada\\
       }


\maketitle

\begin{abstract}
Unsupervised summarization is a powerful technique that enables training summarizing models without requiring labeled datasets. This survey covers different recent techniques and models used for unsupervised summarization. We cover extractive, abstractive, and hybrid models and strategies used to achieve unsupervised summarization. While the main focus of this survey is on recent research, we also cover some of the important previous research. We additionally introduce a taxonomy, classifying different research based on their approach to unsupervised training. Finally, we discuss the current approaches and mention some datasets and evaluation methods.
\end{abstract}

\section{Introduction}
\label{Introduction}

Text summarization aims to generate a summary that is concise, fluent, and covers the main points from the input. With the growing amount of textual content on the internet, such as news, social media, and reviews, it is inefficient to have humans summarize large amounts of information. Natural language processing techniques offer an ideal alternative as they already have shown great success in areas such as translation \shortcite{t5-DBLP:journals/corr/abs-1910-10683}. However, in addition to the technical challenges of natural language understanding and natural language generation, text summarization techniques often require large amounts of labeled training data in form of source and summary pairs. While this training data may be readily available in specific domains, such as news, curating labeled data in other domains and low-resource languages is often expensive. Furthermore, languages evolve, and new domains are introduced, making creating datasets for every domain and keeping them up to date virtually impossible. Unsupervised algorithms aim to overcome mentioned issues by not relying on labeled datasets for training; instead, they use techniques that enable learning from unlabeled data. As a result, unsupervised models can be trained on unlabeled datasets, often much larger than labeled ones, and achieve state-of-the-art performance in their respective fields \shortcite{bert-DBLP:journals/corr/abs-1810-04805,radford2019language}.

This survey mainly focuses on categorizing different approaches to unsupervised summarization techniques in recent years. In the past few years, there has been a significant amount of research on the topic; however, to the best of our knowledge, there has yet to be a survey
categorizing
the different approaches to unsupervised summarization\shortcite{10.1145/3529754,carichon2023history}. Current summarization research often classifies based on the number of inputs, single document summarization (SDS) and multi-document summarization (MDS), the focus of summary, generic or aspect-based, or the summarization strategy, abstractive and extractive. Extractive methods find and select salient words or sentences from the input document and present them as the summary, usually by ranking or selection. Abstractive methods generate the summary based on their understanding of the input document, usually by encoding and decoding the input document into the summary. However, these classifications are broad and do not seem to convey the details of the ideas and strategies used for summarization. We 
propose a
fine-grained classification based on the type of techniques and methods recently employed for the task.  
We first break the methods into three categories, abstractive, extractive, and hybrid; then, methods in each category are classified based on their specific approach.
In addition to unsupervised methods, we also cover weakly-supervised, self-supervised, zero-shot, and few-shot methods since they tackle the same problem, lack of labeled data. Weakly-supervised methods use noisy data, and self-supervised methods use auxiliary tasks for generating training signals. Semi-supervised learning, a subcategory of weakly-supervised learning, uses a small labeled dataset and a larger unlabeled dataset in combination.


 \begin{itemize}
    \item We propose a taxonomy for classifying unsupervised methods in summarization.
    \item We cover recent approaches and research in unsupervised summarization.
    \item We discuss the current trends, and limitations of summarization methods.
\end{itemize}

The structure of this survey is as follows. In section 2, we present our own taxonomy of recent research in unsupervised summarization and explain different approaches used. Section 3 covers abstractive methods. Section 4 reviews extractive methods. Section 5 explores hybrid methods. Section 6 discusses the problems and limitations of the approaches we covered. Lastly, section 7 mentions popular datasets in different domains and the evaluation methods used. Our contributions are as follows:

\section{Taxonomy}
\label{network}

This section covers how we categorize methods in unsupervised summarization and the motivation behind it. We distinguish \textbf{extractive} and \textbf{abstractive}, mainly because abstractive and extractive methods are usually different in their techniques and models. 

We further break down \textbf{abstractive models} into language model-based methods, methods that use reconstruction networks, and other methods. The \textbf{language model-based methods} category contains recent research that uses pretrained language models or large language models for summary generation. The \textbf{reconstruction networks} consist of methods that try to train a model with reconstruction as one of their objective functions. This category mainly contains abstractive methods prior to the introduction of large pretrained language model. The last category of abstractive models includes \textbf{other methods} that could not be classified into the other two categories. 

\textbf{Extractive methods} are categorized into three sub-classes, selection, ranking, and search-based. \textbf{Selection methods} perform binary classification on each unit (word, phrase, sentence, or document) to decide whether they are salient enough to be included in the summary. \textit{Ranking methods} use different techniques to score and rank units and choose the
\begin{figure}[htp]
    \centering
    \includegraphics[width=6cm]{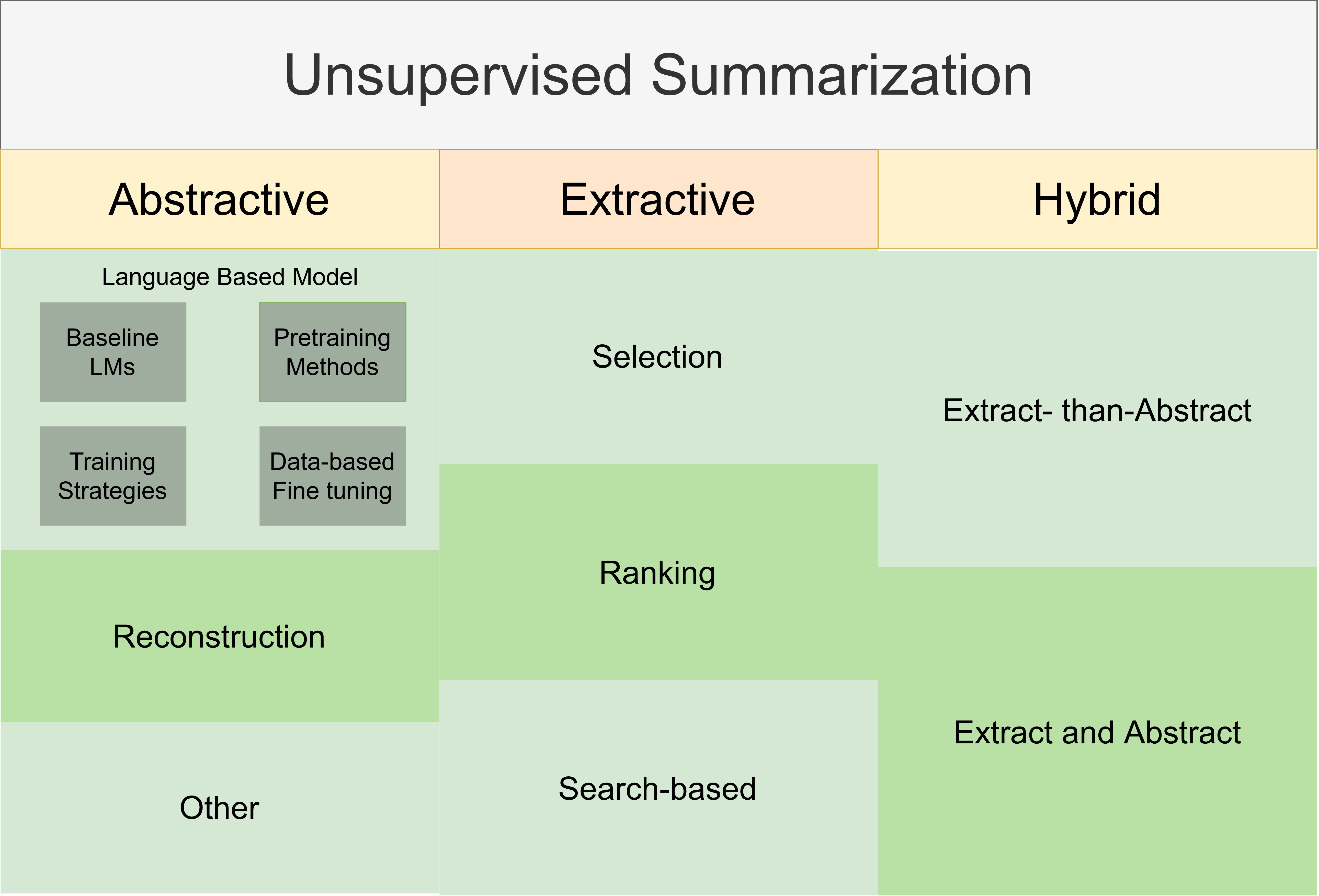}
    \caption{Hierarchical Structure of Our Taxonomy}
    \label{fig:taxa}
\end{figure}
best ones as the summary. \textbf{Search-based methods} aim to iteratively improve the summary by editing and evaluating the new summary. 

Lastly, we introduce another category called \textbf{hybrid methods}, which use extractive and abstractive summarization techniques. Methods in this section either use both extractive and abstractive techniques for summarization; or introduce a training strategy that can be applied to both extractive and abstractive methods. Figure ~\ref{fig:taxa} outlines the proposed structure.

\section{Abstractive Methods}
\subsection{Language Models}
Pretrained language models (PLMs) and large language models (LLMs) have recently transformed the landscape of NLP and impacted almost every NLP task including text summarization. Both models are transformer-based \shortcite{transformer} models that are pre-trained on large corpora for various NLP tasks. Similar to other natural language generation and natural language understanding tasks, summarization can significantly benefit from models pretrained on a large corpus of data. In fact, these models often generate satisfactory summaries even without any fine-tuning or task-specific training; as a result, they are often used as baseliens in recent research \shortcite{zou-etal-2020-pre,xiao-etal-2022-primera}. Their performance can further be improved using few-shot training with as few as ten training examples as shown by \shortcite{goodwin-etal-2020-flight}. Abstractive summarization aims to generate a summary based on its understanding of the input text. This enables both encoder-decoder architecture and decoder-only architecture to be applicable for the task. While the encoder-decoder architecture was more popular prior to 2023, the introduction and the following popularity of ChatGPT shifted the focus in research towards decoder-only LLMs \ref{table:1}. Moreover, the research utilizing encoder-decoder PLMs often centered around trained and fine-tuned the models with various strategies in order to improve their summarization capabilities; whereas the recent LLM-based research often focus on prompt-engineering instead of fine-tuning. This may be a result of computational power required for fine-tuning LLMs and availability of models and their weights to the public.

This section covers \textbf{Baseline Language Models}, \textbf{Pretraining Methods} for language models, \textbf{Training Strategies} for language models, and \textbf{Fine-tuning} PLMs for summary generation.

\subsubsection{Baseline Language Models}
Baseline language models section focuses on research done utilizing Pretrained Language Models (PLMs) and Large Language Model (LLMs) without applying any training or fine-tuning. These work discuss the performance of LMs in either zero-shot or few-shot setting, without modifying the model's weights.

In our categorization, we refer to encoder-decoder models such as BART \shortcite{lewis-etal-2020-bart} as PLMs, and larger, decoder-only models such as GPT-4 \cite{openai2024gpt4} as LLMs. Although this categorization is not completely accurate as LLMs are pretrained models and PLMs are also considered as large language models, we selected this terminology for more convenient reference to these models.

\smallskip
\textbf{3.1.1.1 Pretrained Language Models} \smallskip

BART \shortcite{lewis-etal-2020-bart}, T5 \shortcite{t5-DBLP:journals/corr/abs-1910-10683}, and PEGASUS \shortcite{pegasus-10.5555/3524938.3525989} are three of most popular encoder-decoder PLMs utilized in summarization task. They show strong performance in zero-shot, few-shot, and fully fine-tuned settings \shortcite{goodwin-etal-2020-flight}, and are often used as baselines for performance evaluation \shortciteA{xiao-etal-2022-primera,yu-etal-2021-adaptsum}. 
According to \shortciteA{fabbri-etal-2021-improving}, in some case summaries generated by these models were favored over human-written ones. 
BART and T5 are general-purpose pre-trained language models. The former is trained on document rotation, sentence permutation, text-infilling, and token masking and deletion objectives, while the latter is trained on token masking, translation, classification, reading comprehension, and summarization. Contrastively, PEGASUS was explicitly trained for summarization using self-supervised masked language modeling and gap-sentence-generation (GSG) objective \shortcite{pegasus-10.5555/3524938.3525989}. In GSG, the model has to predict the masked sentence conditioned on the other sentences in the input, where the masked sentence is the a salient sentence resembling the summary. 

Regarding the performance of these PLMs, the authors of PEGASUS report that PEGASUS outperforms both BART and T5 on certain datasets. \shortciteA{goodwin-etal-2020-flight} shows that the performance of the three models is dependant on the dataset and reports both T5 and BART outperforming PEGASUS on some datasets in zero-shot setting. As for few-shot setting, PEGASUS showed significant improvements compared to other two models and was the more consistent model across various datasets on average \shortcite{goodwin-etal-2020-flight}.

\smallskip
\textbf{3.1.1.2 Large Language Models} 
\smallskip

The recent emergence and success of GPT-based models by OpenAI has made them a popular choice for many NLP tasks, including summarization. ChatGPT \cite{ouyang2022training} is a large language model, with considerably more parameters, trained for next word prediction and fine-tuned with reinforcement learning from human feedback. Due to their recency, LLMs such as ChatGPT, GPT-4, and LLAMA-2 \cite{touvron2023llama} are rather understudied for downstream NLP tasks, e.g. summarization, however recent works aim to examine the capabilities of LLMs. Works such as \shortcite{wang-etal-2023-element}, \shortcite{syed-etal-2023-indicative}, \shortcite{adams-etal-2023-sparse}, \shortcite{tang-etal-2023-context}, and \shortciteA{wang-etal-2023-zero} focus on prompt engineering and utilizing LLMs in zero-shot and few-shot (also called in-context learning) setting, and/or aim to compare the evaluate and compare performance of various LLMs.

\shortciteA{wang-etal-2023-element} \textbf{examine reference summaries} of two popular summarization datasets, CNN/DailyMail \shortcite{cnn-dailymail/DBLP:journals/corr/HermannKGEKSB15} and BBC XSum \shortcite{narayan-etal-2018-dont}, and compare the performance of pre-trained language models, BART, T5, and PEGASUS, to large language models, GPT-3. Authors aim to address the current reference summaries issues (e.g. factual hallucination and information redundancy), by \textbf{writing new reference summaries} based on “Lasswell Communication Model” \shortcite{lasswell}. They further show that zero-shot large language models generate on par, and even better outputs than pre-trained language models, according to automatic evaluation metrics (ROUGE and BERTScore) with the new reference summaries. Authors also utilize a two-step \textbf{Chain-of-Thought prompting} method in GPT-3  that first extracts the important information and generates the summary from extracted information. 

\shortciteA{syed-etal-2023-indicative} and \shortciteA{adams-etal-2023-sparse} explore the ideal structure and properties of a reference summary. \shortciteA{syed-etal-2023-indicative} presents an unsupervised approach for summarizing long discussions on online forums using various LLMs. The authors suggest generating an "indicative" summary of long discussions, where only the gist of different aspects in topic are presented as the summary (similar to a book’s table of content); as opposed to "informative" summaries that aim to capture as much information as possible. The authors use a three step approach for creating the summary. First, the sentences of all the arguments are clustered. Second, an LLM is used to generate a single line summary of each cluster, called a label. Finally, the labels are assigned to one or more frames, where the frames are different aspects of a topic (e.g. a topic about a political idea can have aspects such as economy or ideology). The experiments show that while open-source LLMs (LLAMA and T0) have acceptable results, they are outperformed by OpenAI’s GPT models (ChatGPT and GPT4). 
\shortciteA{adams-etal-2023-sparse} analyze GPT-4 generated summaries and the optimal amount of "density" for summaries. The authors argue that an ideal summary should be entity-centric without being too difficult to understand, and the level of entities included in a summary determines its density. The authors propose a technique called “Chain of Density Prompting”, that enables iteratively prompting a LLM to generate more dense summaries. Their experiments show that human written summaries have a higher density compared to vanilla GPT-4 summaries, and that the humans prefer summaries with densities closer to human written summaries. 

\shortciteA{tang-etal-2023-context} and \shortciteA{wang-etal-2023-zero} benchmark the performance of various LLMs in specific summarization tasks. The work of \shortciteA{tang-etal-2023-context} focuses on analyzing the performance of various LLMs in dialogue summarization. Specifically, the authors conduct various experiments for generating both controlled (e.g. length control) and uncontrolled summaries using in-context or few-shot learning. Their experiments show that LLMs perform reasonably well, and models such as LLAMA and Alpaca achieve high factual consistency scores. \shortciteA{wang-etal-2023-zero} focus on cross-lingual summarization, i.e. summarizing input text in another language, by comparing the performance of different LLMs in zero-shot setting. Their experiments showcase that ChatGPT and GPT-4 perform well and produce detailed summaries. Furthermore, they report that GPT-4’s performance in zero-shot setting is comparable to an mBART-50 \shortcite{tang-etal-2021-multilingual} fine-tuned for the task. 

\subsubsection{Pretraining Methods}

\shortciteA{zou-etal-2020-pre} and \shortciteA{he-etal-2023-z} focus on introducing more efficient training objectives for improving the performance of pretrained language models, enabling them to outperform larger pretrained language models that are trained on more data. In \shortciteA{zou-etal-2020-pre}, the authors introduce three unsupervised pretraining objectives for training sequence-to-sequence based models. They argue that the proposed unsupervised objectives, sentence reordering, next sentence generation, and masked document generation, are closely related to the summarization task, resulting in improved abstractive summaries. Using RoBERTa \shortcite{roberta-DBLP:journals/corr/abs-1907-11692} as the encoder, they show that their model achieves comparable results with only 19 GB of data compared to PLMs trained on more than 160 GB of text. 
\shortciteA{he-etal-2023-z} introduces a new encoder-decoder pre-trained language model for summarization. The authors similarly propose three techniques for improving the pre-training stage. First, the model is pre-trained on replaced token detection and corrupted span prediction, and is then on trained document-summary pairs for summary generation. The second technique replaces the self-attention layers in the encoder with the disentangled attention layer. Disentangled attention represents each token with two vectors, content and position, to improve the effectiveness of the model. Third, they use the fusion-in-encoder for handling long sequences.

\shortciteA{xiao-etal-2022-primera} and \shortciteA{pagnoni-etal-2023-socratic} introduce pretraining objectives for specific tasks and goals. \shortciteA{xiao-etal-2022-primera} focus on an objective function for multi-document summarization, utilizing the gap sentence generation objective. They improve upon the gap sentence generation objective by clustering related sentences together, and masking the most informative sentence in the cluster, using the rest of the sentences in that cluster to predict the masked sentence. After applying their pretraining method on Longformer-Encoder-Decoder \shortcite{led-DBLP:journals/corr/abs-2004-05150}, they report noticeable improvements in zero-shot setting and marginal improvements after supervised training. 
\shortciteA{pagnoni-etal-2023-socratic} introduces an unsupervised pre-training objective to improve controllability of summaries. In this method, the summarization model is asked to answer questions that are automatically generated based on the input document, which allows the method to make use of unlabeled documents. Specifically, the method first selects important sentences from the input document as pseudo-summaries using ROUGE, and generates questions about them using MixQG \shortcite{murakhovska-etal-2022-mixqg}. The model is trained to generate the questions and answer them given the document. This ensures that the generated summary accurately addresses the user's query by focusing on pertinent content.

\shortciteA{yu-etal-2021-adaptsum} and \shortciteA{chen-etal-2023-unisumm} experiment on utilizing pretaining objective for PLMs in areas with not enough training data. \shortciteA{yu-etal-2021-adaptsum} experiments on domain adaptation for low-resource summarization tasks such as email summarization. They experiment on further pretraining BART using three different training objectives, source domain pretraining, domain-adaptive pretraining, and task-adaptive pretraining. In source domain pretraining the model is pretrained on labeled data from the source domain (i.e. any domain with substantial labeled data that is not our target). In domain-adaptive pretraining, the model is pretrained on an unlabeled domain-related corpus (i.e. documents in the target domain). Lastly, in task-adaptive pretraining, the model is pre-trained on a smaller unlabeled domain-related corpus that is more task-relevant. Their experiments show that source domain pretraining and task-adaptive pretraining can generally improve performance, whereas the effectiveness of domain-adaptive pretraining depends on the pretraining data and the target domain task data. 
\shortciteA{chen-etal-2023-unisumm} proposes a new few-shot summarization model pre-trained for different summarization tasks on different datasets. This enables leveraging the shared knowledge available in different datasets. To accomplish this the authors use prefix-tuning \shortcite{li-liang-2021-prefix}, where the main idea is to extract knowledge prepending and tuning additional parameters, called prefixes, before each layer of the PLM. In their approach, the authors first pre-train a summarization model with task-specific prefix vectors using multi-task pre-training. During inference on a new task, the prefix vector is fine-tuned in a few-shot setting from a universal prefix that was trained in the pre-training stage. Their experiments shows that their approach outperforms baselines and achieves comparable results to GPT-3.5.

\subsubsection{Training Strategies}
The models in this section employ different unsupervised training signals to fine-tune existing language models. The goal of these strategies is to improve the fluency, coverage, or factual consistency of generated summaries. 

\shortciteA{oved-levy-2021-pass} focus on review summarization, arguing that previous models suffer from two problems, repetitive and factual inconsistent summaries. As a solution, they first fine-tune T5 on a small labeled review summarization model. In order to produce summaries, the model generates multiple candidate summaries for each subset of reviews. A subset is created by dropping k reviews out of the input and concatenating the rest, to increase diversity of generated candidates. Next, using human annotated coherence scores, they train a "coherence summary ranker" model to score the candidate summaries with respect to their coherence and factual consistency. The candidate with the highest score is selected as the summary. The experiments show that the proposed approach generates informative summaries while being diverse and coherent.

\shortciteA{laban-etal-2020-summary} and \shortciteA{roit-etal-2023-factually} propose new training signals to fine-tune language models using reinforcement learning. \shortciteA{laban-etal-2020-summary} introduces a novel approach for fine-tuning a language model (i.e. GPT-2) in order to optimize the fluency and coverage of keywords in the summary while enforcing a length limit as a result of model's decoder-only architecture. The authors combine two scores, fluency score and coverage score, in order to produce a training signal for the model using the self-critical sequence training (SCST) method \shortcite{scst}. The fluency score is calculated using by GPT-2 using the log-probability of the generated summary. The coverage score is calculated by first masking the keywords in input document, and feeding it alongside the generated summary to a fine-tuned BERT model to predict the masked words. The accuracy of the BERT model is used as the coverage score. The proposed approach outperforms previous unsupervised summarization baselines, and are comparable to supervised approaches at the time in terms of ROUGE score. 
\shortciteA{roit-etal-2023-factually} focus on generating factual consistent summaries using reinforcement learning. They propose using textual entailment of input document and generated summary, as the reward for the reinforcement learning model. The idea behind the approach is that models trained on natural language inference datasets can accurately detect factual inconsistencies. However to ensure the model had summary generation capabilities, i.e. generating coherent and relevant text, it was first pre-trained to produce summaries using maximum-likelihood objective, and only further fine-tuned using reinforcement learning. Their results show improvements over baselines for factual consistency, salience, and conciseness.

\subsubsection{Data based Fine-tuning}
The methods in this section generate or collect a dataset in an unsupervised or weakly supervised manner, for a specific task or domain where there is not enough data available. 

\shortciteA{fabbri-etal-2021-improving} and \shortciteA{tan-etal-2020-summarizing} focus on generating a labeled dataset for summarization using websites and resources available online. \shortciteA{fabbri-etal-2021-improving} introduced an unsupervised method for extracting pseudo-summaries from Wikipedia for fine-tuning PLMs. These pseudo-summaries have similar high-level characteristics to our ideal target summaries, where the high-level characteristics refer to length or the level of abstractiveness. They use the first \textit{k} sentences in each article as the summary and the rest \textit{i} sentences as the input. They further augment the dataset by using round translation \shortcite{roundtr} and is the dataset to fine-tune BART. The experiments show that the fine-tuned BART outperforms the vanilla BART in zero and few-shot setting across various datasets. 
\shortciteA{tan-etal-2020-summarizing} aims to facilitate aspect summarization on any given aspect; however, since it is impossible to create a dataset for every aspect, they employ a weakly supervised approach for data collection. In order to collect the training data different aspects of input are extracted from a labeled summarization dataset using an NER model, and all other related aspects to it are also extracted using ConceptNet \shortcite{conceptnet}. The summary of each aspect, i.e. sentences in reference summary related to the aspect (or similar aspects found in ConceptNet) are concatenated together. During summary generation, the aspect, a list of related words (collected from the Wikipedia page of the aspect), are given as context alongside the input document as a single input to the summarization model. 

\shortciteA{laskar-etal-2020-wsl} and \shortciteA{pham-etal-2023-select} use language models to generate labeled datasets for specific summarization tasks. \shortciteA{laskar-etal-2020-wsl} tackles query-focused summarization in the multi-document (QF-MDS) setting, where the goal is to generate a summary from multiple documents given a query. Due to the absence of specific training data for the task, the authors adopt a supervised training approach utilizing other QF-MDS datasets as substitutes and generate weak reference summaries from reference summaries. In the next step, the weak reference summaries are used to fine-tune BERTSUM model \shortcite{bertsum-DBLP:journals/corr/abs-1908-08345} to generate the query focused abstractive summaries. Finally, the generated summaries are ranked by their relevance to the query and most relevant ones are selected as the summary.
\shortciteA{pham-etal-2023-select} proposes an approach for generating weakly-supervised training data using LLMs. The authors argue that fine-tuning LLMs for specific tasks is infeasible due to the resources required, and collecting enough training data for fine-tuning smaller language models such as BART in order to match their performance to LLMs is expensive. Therefore, the authors propose a three step knowledge distillation technique for creating training data using LLMs. The approach generates annotated data from a large corpus of unlabeled data, with the help of a small validation set. Specifically, the method first extracts text from the large unlabeled corpus that is semantically similar to text in the validation set. Second, it creates a prompt with the labeled data from validation set as few-shot examples, and uses ChatGPT to predict the summary for the sampled text using the prompt as input. In the final step it filters low-quality summaries generated by ChatGPT. Their results show improvements over standard knowledge distillation approaches.

Overall, methods that use PLMs or LLMs achieve better performance on average compared to other methods in this survey. However, they still do suffer from the common problems in abstractive summarization, such as factual correctness \shortcite{oved-levy-2021-pass}. Additionally, training or fine-tuning such models poses limitations on resource and introduces new challenges. For example, \shortciteA{yu-etal-2021-adaptsum} have shown that continues pretraining causes catastrophic forgetting, where the pretrained language model loses some of its language understanding ability gained in the initial pretraining step.



%




\subsection{Reconstruction}
The goal of reconstruction approach is to reconstruct the original input from the modified input. As a result it is a popular training strategy for unsupervised summarization since the original document can be used as the reference summary (output), while the source (input document) can be created by adding noise to the original document. Models trained with reconstruction loss often use an encoder-decoder architecture at their core, where they first try to encode the input document into a representation and then decode the representation to the summary. Ideally, the encoder should encode the salient information from the input, and the decoder should decode the representation into a fluent text that captures the vital information and is shorter than the input. In order to fulfill these requirements, summarization models use different techniques.

\shortciteA{meansum-DBLP:journals/corr/abs-1810-05739}, \shortciteA{baziotis-etal-2019-seq}, and \shortciteA{CARICHON2023109839} use auto-encoder architecture as part of their approach for unsupervised summarization. \shortciteA{meansum-DBLP:journals/corr/abs-1810-05739} uses an auto-encoder trained for reconstruction for creating representations, \shortciteA{baziotis-etal-2019-seq} uses two auto-encoders one for compression and the other for reconstruction, and \shortciteA{CARICHON2023109839} trains two auto-encoders for language modelling and constraining length.

\shortciteA{baziotis-etal-2019-seq} proposes a novel architecture for summarization emphasizing on sentence compression using two attention-based encoder-decoder. The first encoder-decoder is the compressor, tasked with summarizing the input, and the second encoder-decoder reconstructs the original input from the summary. In order to train the model, four losses are computed. The first loss is the reconstruction loss between the original input and the output of the second decoder; the second is the language model prior loss to ensure the generated summary is fluent. The third loss, topic loss, encourages coverage by calculating the cosine distance between the average of word embeddings in input and summary. Lastly, the fourth loss enforces the length penalty.
\shortciteA{meansum-DBLP:journals/corr/abs-1810-05739} proposes a method for multi-document summarization using an LSTM-based encoder-decoder. The model is composed of two parts; the first part is an encoder-decoder that tries to learn the representation for each input using reconstruction loss. The second part learns to generate a summary similar to each of the inputs. The second part is trained using the average similarity between the representation of each input and the representation of the summary generated from the mean of input representations. 
\shortciteA{CARICHON2023109839} focus on “update summarization” of news by iteratively updating a summary as new information is available in social media settings using autoencoders. The proposed autoencoder model is trained for two tasks simultaneously, language modeling and information constraint which is implemented by limiting the length of the text. The proposed approach, unlike previous extractive approaches, does not rely on classifying redundant text to find salient information. This is important as social media posts are concise and not repetitive as opposed to traditional news stories, which renders methods that rely on classification of redundant information useless. The proposed approach shows improvements in ROUGE scores over previous baselines.

\shortciteA{amplayo-lapata-2020-unsupervised} focus on unsupervised opinion summarization, and creating a synthetic dataset by introducing noise. They introduce two linguistically motivated noise generation functions to add noise, and train the model to denoise the input. The noise generation functions add noise on word/phrase level (e.g., changing words) and on document level (replacing a review with a similar one). For the model, they use an LSTM-based encoder-decoder with attention \shortcite{attention} and copy \shortcite{vinyals-copy} mechanisms in the decoder which sets it apart from the previous auto-encoder based models. They experiment on movie and business datasets and show improvements over previous baselines.

Before the language models, reconstruction networks such as \shortciteA{baziotis-etal-2019-seq} used to be the state-of-the-art for abstractive summarization. However, because they are trained to reconstruct input instead of summary, they cannot learn the key characteristics of summaries \shortcite{brazinskas-etal-2020-shot}. Additionally, due to the emphasis of methods on conciseness, they tend to mix generic statements with informative content \shortcite{oved-levy-2021-pass}. Lastly, reconstruction networks that use autoencoders are limited to simple decoders, lacking attention and copy mechanism that has proven to be useful in summary generation \shortcite{amplayo-lapata-2020-unsupervised}.




\subsection{Other}
The last classification of abstractive summary includes approaches that often utilize novel techniques for summarization. The works in this section use non-autoregressive architecture, custom networks, and reinforcement learning techniques as part of their approach. 

\shortciteA{liu-etal-2022-learning}, propose a non-autoregressive method for sentence summarization. They use the method introduced by \shortciteA{schumann-etal-2020-discrete} to generate summaries of news articles and use this data to train an encoder-only transformer using Connectionist Temporal Classification (CTC) \shortcite{ctc} algorithm for summarization. The non-autoregressive architecture enables length controlling, which can be desirable in summarization tasks. \shortciteA{liu-etal-2022-learning} claims their non-autoregressive approach to summarization using encoder-only architecture is several times faster, and better captures the input–output correspondence compared to autoregressive models. However, the performance of the proposed approach in terms of ROUGE score is not on par with the state-of-the-art.

\shortciteA{brazinskas-etal-2020-shot} introduce a few-shot method for review summarization. They argue that previous unsupervised methods trained on review datasets have not been exposed to actual summaries, therefore the summaries generated by them lack essential properties (e.g.  writing style, informativeness, fluency, etc). To solve this, they aim to train a model to generate a summary based on reviews. Specifically, they first train an encoder-decoder transformer \shortcite{transformer} for the review generation task on a large review dataset, using leave-one-out objective \shortcite{Besag1975StatisticalAO}. During training, the model is conditioned on the properties of reviews, such as writing style, informativeness, fluency, and sentiment preservation, however for the inference phase, when the goal is to generate summaries, not reviews, these properties are predicted by a small plug-in network trained on a handful of reviews with summaries. As a result, we can have a small model trained with a few data samples, guiding our encoder-decoder model that generates the summaries. Their experiments showed improvements over previous SOTA at the time, in both human and automatic evaluation.

\shortciteA{kohita-etal-2020-q} proposed an edit-based approach using Q-learning. Edit-based approaches are often extractive methods that start from a pseudo summary (e.g. a random selection words, or the original text) and perform continues edits to improve the summary. The edits differ depending on the work, but they often include adding words and removing them. After each edit the quality of the summary is evaluated using a scoring function to check if the edit has improved the summary or not. 
In \shortciteA{kohita-etal-2020-q}, the method consists of an agent that selects an action (keep, delete, or replace), and a language model applies the edits to generate summary. The agent predicts an action, delete, keep, or replace, for each word; the first action deletes the word, the second keeps it as is, and the third action replaces the word with a [MASK] token. The new sentence is given to BERT in order to predict masked tokens and generate the summary. In the training phase, after the summary is generated, it is further reconstructed back to the original in order to calculate the step reward score, similar to reconstruction methods. The step reward score combined with summary assessment, which evaluates informativeness, shortness, and fluency, is the reward that is used for calculating the loss function. Their experiments report on-par results compared to previous baselines.


\section{Extractive Methods}
Extractive methods create a summary by selecting salient units (words, phrases, or sentences) from the input document. Extractive methods often view summarization tasks either as a sequence selection task, where units are classified as to be included in the summary or not, or use a ranking technique to rank all the units by their salience and select the top k as the summary. A less researched approach is search-based summarization, where the goal is to maximize an objective function that evaluates the summary by editing it. These edits are usually composed of add, delete, and replace, and the search space is the units in the input document.

\subsection{Selection}

\shortciteA{zhang-etal-2019-hibert} focus on proposing an unsupervised pretraining approach for training an extractive summarization model. They argue that since large amounts of labeled data for extractive summarization is not available, using an unsupervised pretraining approach enables the model to learn from unlabeled data. Similar to how BERT learns the representation of sentences by predicting words, the authors propose an approach to learn the  representation of documents by predicting sentences. Specifically, they use a hierarchical encoder for the sentence selection. The hierarchical architecture consists of two identical encoders, one for sentence-level encoding with words as inputs and the other for document-level encoding with sentence encodings as inputs and one decoder. For the training objective, masked language modeling is used where every word in 80\% of sentences is masked, and the model is tasked with predicting the masked tokens. For summary generation, the output of the second encoder is used to predict a true or false label for each sentence in the document. Their experiments show some improvements in ROUGE compared to baselines. 

\shortciteA{wang-etal-2022-noise} proposes a semi-supervised approach using consistency training. Consistency training aims to make the model resilient to slight changes by introducing noise and tasking the model to generate consistent summaries. To implement this, authors inject noise by replacing words with semantically similar words using BERT and train the model to generate similar summaries from original and noisy inputs. However, training using consitency training is not enough by itself, as there is not enough labeled data for the task. To this end, the authors also propose an entropy-constrained pseudo-labeling strategy that is used for getting high-confidence labels from unlabeled data. The pseudo label is assigned by comparing the entropy of the predicted result of unlabeled data and the predicted result of labeled data, where the pseudo label is preserved if the entropy of unlabeled data is smaller than the entropy of labeled data. Their experiments show slight improvements over baselines.

\shortciteA{basu-roy-chowdhury-etal-2023-unsupervised} proposes a two-step opinion summarization method based on representation learning and a geodesic distance-based selection function. The proposed approach first generates topical representations of input sentences using dictionary learning. Next it uses its selection function to select sentences as the summary, based on the geodesic distance between their representations. The goal of the approach is to extract sentences from opinions that are shared between users as it shows their importance. The authors claim that the learned topical representations using dictionary learning better capture the semantics of a text, compared to pre-trained models. Moreover, they use a geodesic distance-based function for computing the importance of a review, as it considers the “underlying manifold” of representations. Their experiments showcase that their method outperforms previous unsupervised opinion summarization methods.

In addition to the common problems of extractive summarization, selection methods tend to keep the relative order of sentences in the input, which is a limitation \shortcite{zou-etal-2020-pre}. Also, generating datasets for these models can be difficult since they require sentence-level labels \shortcite{zhang-etal-2019-hibert}.

\subsection{Ranking}
\shortciteA{zheng-lapata-2019-sentence} improve upon a popular graph-based baseline, TextRank \shortcite{mihalcea-tarau-2004-textrank}. TextRank first creates a graph from input, where sentences are the nodes, and the edges are the similarity scores between them. Next, the centrality of each node (i.e. sentence) is calculated, the nodes are ranked by their centrality and the top ones are chosen as the summary. The authors improve upon this by making two changes. First, they used BERT for generating sentence representations, improving the accuracy of sentence similarities, and second, they made the edges in the graph directed. This allows taking the relative position of sentences with respect to each other into account, which enables prioritizing earlier sentences in a document that are more general. Their experiments show signifcant improvement over original TextRank but perform slightly worse than previous SOTA. 

\shortciteA{liang-etal-2022-efficient} focus on summarizing long documents by introducing a two-step ranking method using semantic blocks. Semantic blocks are consecutive sentences in a document that describe the same facet. To achieve this, proposed method first finds all the semantic blocks in the input and filters insignificant facets using a centrality estimator. In the next stage, relevant sentences to facets in each block are selected as candidates, and the final summary is selected from candidates using a sentence-level centrality-based estimator. The authors use the previously mentioned approach \shortcite{zheng-lapata-2019-sentence} as the centrality-based estimator. The proposed approach has a similar performance in terms of ROUGE compared to previous SOTA, however the authors report significant improvements in inference speed.

\shortciteA{basu-roy-chowdhury-etal-2022-unsupervised} uses dictionary learning \shortcite{dictionary-learning} and builds upon the work of \shortciteA{angelidis-etal-2021-extractive} which introduced Quantized Transformer (QT) for review summarization. Dictionary learning or sparse coding aims to find a sparse representation of the input data over latent semantic units. The authors use dictionary learning to improve upon the previous work by representing sentences as a distribution over latent units (as opposed to a single latent representation), and aspect-focused summaries. In their approach, the sentences in reviews are the inputs, and the representations are learned by training an encoder-decoder transformer on reconstruction. In summary generation, the mean representation of all the review sentences is calculated, then the relevance score between each sentence and the mean representation is computed, and finally, the top k sentences with the highest score are selected as the summary. Their experiments show improvements over the previous extractive SOTA including QT.

While ranking methods enable extraction of the most salient parts of the input, resulting in a high coverage score, the output summary they generate lacks coherence and cohesion since all the selected sentences are concatenated together.

\subsection{Search-based}
\shortciteA{schumann-etal-2020-discrete} propose a hill-climbing search-based method for sentence summarization that iteratively improves the summary. They initialize by selecting k random words from the input with order intact, with k being the desired length. At each time step, a new sentence summary is sampled by randomly removing a word summary and selecting another from the original sentence while preserving the relative order of words. The new sentence summary is selected if it achieves a higher score computed by an objective function that evaluates fluency (using language model perplexity) and similarity score between the summary and original sentence.
As mentioned by \shortciteA{liu-etal-2022-learning}, the search-based methods are slow at inference as the methods need hundreds of search steps for each input sentence. These methods are also often restricted to keep the same word order as the input which affects their coherence.

\begin{table*}[t]
\centering
\resizebox{\textwidth}{!}{\begin{tabular}{c|ccm{10em}}
& Works & Method & Pros and Cons\\[0.5ex] 
\hline
\multirow{28}{*}{Abstractive}
& \shortciteA{lewis-etal-2020-bart}    & PLM     & \multirow{28}{*}
{\begin{tabular}[c]{@{}l@{}}
Pros: \\
High performance, \\
Ability to understand \\
and generate new \\
words/sentences \\

Cons: \\ 
High training \\
and inference cost, \\ 
Hallucination, \\
Text degeneration \\
Topic drift, \\
Factual correctness \\
\end{tabular}} \\
& \shortciteA{t5-DBLP:journals/corr/abs-1910-10683}   & PLM     & \\
& \shortciteA{pegasus-10.5555/3524938.3525989}    & PLM     & \\
& \shortciteA{wang-etal-2023-element}    & LLM     & \\
& \shortciteA{adams-etal-2023-sparse}    & LLM     & \\
& \shortciteA{tang-etal-2023-context}    & LLM     & \\
& \shortciteA{wang-etal-2023-zero}    & LLM     & \\
& \shortciteA{syed-etal-2023-indicative}    & LLM     & \\
& \shortciteA{zou-etal-2020-pre}    & LM:PM     & \\
& \shortciteA{xiao-etal-2022-primera}    & LM:PM     & \\
& \shortciteA{yu-etal-2021-adaptsum}    & LM:PM     & \\
& \shortciteA{chen-etal-2023-unisumm}    & LM:PM     & \\
& \shortciteA{pagnoni-etal-2023-socratic}    & LM:PM     & \\
& \shortciteA{he-etal-2023-z}    & LM:PM     & \\
& \shortciteA{oved-levy-2021-pass}    & LM:TS     & \\
& \shortciteA{laban-etal-2020-summary}    & LM:TS     & \\
& \shortciteA{roit-etal-2023-factually}    & LM:TS     & \\
& \shortciteA{fabbri-etal-2021-improving}    & LM:DFT     & \\
& \shortciteA{tan-etal-2020-summarizing}    & LM:DFT     & \\
& \shortciteA{laskar-etal-2020-wsl}    & LM:DFT     & \\
& \shortciteA{pham-etal-2023-select}    & LM:DFT     & \\

& \shortciteA{amplayo-lapata-2020-unsupervised}    & Reconstruction     & \\
& \shortciteA{baziotis-etal-2019-seq}    & Reconstruction     & \\
& \shortciteA{meansum-DBLP:journals/corr/abs-1810-05739}    & Reconstruction     & \\
& \shortciteA{CARICHON2023109839}    & Reconstruction     & \\
& \shortciteA{brazinskas-etal-2020-shot}    & Other     & \\
& \shortciteA{liu-etal-2022-learning}    & Other     & \\
& \shortciteA{kohita-etal-2020-q}    & Other     & \\
 \hline
\multirow{6}{*}{Extractive}  
& \shortciteA{zhang-etal-2019-hibert}    & Selection     & \multirow{6}{*}
{\begin{tabular}[c]{@{}l@{}}
Pros: \\
Faster training \\
and inference, \\
Cons: \\
Worse performance \\
Fluency and coherence \\
\end{tabular}}\\
& \shortciteA{wang-etal-2022-noise}    & Selection     &\\
& \shortciteA{basu-roy-chowdhury-etal-2023-unsupervised}    & Selection & \\
& \shortciteA{zheng-lapata-2019-sentence}    & Ranking     &\\
& \shortciteA{liang-etal-2022-efficient}    & Ranking     &\\
& \shortciteA{basu-roy-chowdhury-etal-2022-unsupervised}    & Ranking     &\\
& \shortciteA{schumann-etal-2020-discrete}    & Search-based     &\\

 \hline
\multirow{5}{*}{Hybrid}      
& \shortciteA{suhara-etal-2020-opiniondigest}    & Ext-than-Abs     & \multirow{4}{*}{\begin{tabular}[c]{@{}l@{}}
Pros: \\
Scalable, \\
Modular \\ 
Cons: \\
Complex architecture
\end{tabular}}\\
& \shortciteA{zhou-rush-2019-simple}    & Ext\&Abs     &     \\
& \shortciteA{fu-etal-2021-repsum}    & Ext\&Abs     &     \\
& \shortciteA{west-etal-2019-bottlesum}     & Ext\&Abs     &  \\
& \shortciteA{hosking-etal-2023-attributable}    & Ext\&Abs     & 
\end{tabular}}
\caption{Categorizing works in unsupervised summarization by their approach.}
\label{table:1}
\end{table*}

\section{Hybrid Methods}

As the name suggests, hybrid models use both extractive and abstractive models in their summary generation process. Some of these methods use an extract-than-abstract method, where an extractive technique is used to select only parts of the input, then an abstractive model is used to generate the summary from the extracted input. The other line of work either focuses on training strategies (rather than extractive or abstractive models), where the training strategy can be applied to both extractive and abstractive models. Or use both extractive and abstractive techinques for summarization. 

\subsection{Extract-than-Abstract}
\shortciteA{suhara-etal-2020-opiniondigest} employ an extract-than-abstract approach for review summarization. Their approach to review summarization consists of three steps: extracting opinion phrases from reviews, selecting desired opinions (based on popularity or specific aspect), and a generator for constructing summaries from selected opinions. During training, they extract opinions from a review using Snippext \shortcite{Snippext-10.1145/3366423.3380144} and train a transformer on generating the original review given opinions. In inference, in order to generate a summary from multiple reviews, first, the opinions are extracted and clustered, similar opinions are merged and the most popular one in each cluster is selected as input to the previously trained transformer to generate a summary. Their approach shows improvements over baselines at the time.  

\subsection{Extractive and Abstractive}
\shortciteA{zhou-rush-2019-simple} propose an approach using two language models for generating a summary that can be used in both abstractive and extractive setting. The model uses the product-of-experts model \shortcite{hinton2002}using two decoder-only language models to predict the next word in the summary by computing the product of their probabilities. The first language model is used for contextual matching (selecting relevant information), while the other is fine-tuned to ensure fluency. This approach could used in both extractive and abstractive setting, since the vocabulary of language models could either be restricted to input document (extractive) or to the vocabulary of language model itself (abstractive). Their approach had competitive results compared to its baselines without any joint training of the models.

\shortciteA{west-etal-2019-bottlesum} uses the information bottleneck \shortcite{bottleneck-2001} principle as a signal for unsupervised summarization to train an extractive and an abstractive model. The information bottleneck aims to produces a summary (x) that is optimized for predicting another relevant information (Y). The authors claim that this approach is more suitable for summarization in contrast to reconstruction loss, since the goal of this method is to retrieve relevant information instead of training to recreate all the information. For this task, X is the summary generated by the model, and Y is the next sentence, as a result the model tries to find the summary that best predicts the next sentence. The authors use this approach to train extractive model and further use the data generated by the model to train an abstractive model in a self-supervised manner. Their extractive model performs slightly better than unsupervised SOTA at the time in terms of ROUGE, however their abstractive method does not show improvements over previous methods. 

\shortciteA{fu-etal-2021-repsum} and \shortciteA{hosking-etal-2023-attributable} propose an unsupervised method for summarization that can produce both extractive or abstractive summaries. \shortciteA{fu-etal-2021-repsum} proposes a self-supervised training strategy for dialogue summarization that can be applied to both abstractive and extractive approaches. The approach uses a dialogue generator model to predict the next utterance of dialogue under two scenarios. In one, they condition the dialogue generator on previous dialogues, and in the other, they condition the dialogue generator on the summary of the previous dialogues. They use the difference between summaries to calculate the loss function. They use the loss to train both abstractive and extractive models. The idea behind this approach is that a good summary should offer a replacement for the original dialogue; as a result, the next utterance of dialogue generate from both previous dialogues, and the summary should be similar. Both approaches perform better than the baselines at the time.
\shortciteA{hosking-etal-2023-attributable} proposes a scalable and controllable approach for opinion summarization based on hierarchical encoding that can be used in both abstractive and extractive settings. The approach first encodes the inputs into a hierarchical space. Next, depending on the setting, it generates summaries either by extracting opinions belonging to the most popular encodings (extractive), or decodes the most popular encodings (abstractive). Their approach provides scalability as it can aggregate encodings of multiple documents, while also allowing for aspect-based summaries due to the hierarchical nature of their encodings. Their results show general improvements across datasets compared to previous SOTA in terms of ROUGE for both extractive and abstractive settings.

The Extract-than-Abstract approach combines the strengths of both methods and covers their weaknesses; the extractive module finds salient pieces of information, and the abstractive module generates a summary from a much smaller input. As a result, the extractive module does not have to deal with the fluency of generated text, and the abstractive module is not conditioned on long texts, which is an issue for attention and RNN-based models. Extract-than-Abstract models perform exceptionally well when in long or multi-document settings. Extractive and abstractive strategies, on the other hand, are more flexible as they can be applied to both techniques, enabling fair comparison between the two methods. However, since these models use external extractive and abstractive models, their performance relies heavily on them. For example, RepSum \shortcite{fu-etal-2021-repsum} relies on a dialogue generation model for training, restricting the summarization performance on the performance of the auxiliary task.


\setlength\tabcolsep{0pt}
\begin{table*}[t]
\centering
\resizebox{\textwidth}{!}{\begin{tabular}{cccccc}
Name & Paper & Domain & Type & Avg Words & Train/Val/Test \\ [0.5ex] 
 \hline
CNN/Dailymail & \shortcite{cnn-dailymail/DBLP:journals/corr/HermannKGEKSB15} & News & Ext & 781/56 & 287,113/13,368/11,490 \\
Gigaword Corpus & \shortcite{rush-etal-2015-neural} & News & Abs & 25/7 & 3,803,957/189,651/1,951 \\
DUC/TAC & \href{https://duc.nist.gov/}{Website} & News & Ext & N/A & N/A \\
XSum & \shortcite{narayan-etal-2018-dont} & News & Abs & 431/20 &  204,045/11,332/11,334  \\
Multi-News &  \shortcite{fabbri-etal-2019-multi} & News & Abs & 2,103/263 & 44,972/5,622/5,622 \\
Newsroom & \shortcite{grusky-etal-2018-newsroom}  & News & Both & 658/26 & 995,041/108,837/108,862  \\
Yelp & \href{https://www.yelp.com/dataset}{Website}  & Review & Abs & N/A & 1,038,184/129,856/12,984  \\
SPACE & \shortcite{angelidis-etal-2021-extractive}  & Review & Abs & N/A & 1,140,000  \\
Amazon & \shortcite{amazondataset-10.1145/2872427.2883037}  & Review & Abs & N/A & 4,750,000  \\
OPOSUM & \shortcite{angelidis-lapata-2018-summarizing}  & Review & Ext & N/A & 359,000 \\
WikiSum & \shortcite{cohen-etal-2021-wikisum}  & Wikipedia & Ext & 1,334/101 & 39,775  \\
TIFU Reddit & \shortcite{kim-etal-2019-abstractive}  & Social Media & Abs & N/A & 42,139+79,740  \\
arXiv & \shortcite{cohan-etal-2018-discourse}  & Academic & Ext & 6,913/293 & 215,913 \\
PubMed & \shortcite{cohan-etal-2018-discourse}  & Academic & Ext & 3,224/214 & 133,215 \\
\end{tabular}}
\caption{List of popular datasets in summarization alongside their domain, reference summary type (extractive or abstractive), average word length for input document and reference summary, and train, validation, test splits or total number of data points}
\label{table:2}
\end{table*}

\section{Discussion and Trends}
\subsection{Supervised and Unsupervised}
While current unsupervised methods offer lower performance in terms of ROUGE than their supervised counterparts \shortcite{kohita-etal-2020-q}, being independent of labeled datasets makes them a desirable approach in many areas. In some cases, an unsupervised approach is the only possible approach, for example, when dealing with a new area with no dataset or frequently changing domains such as social media. Additionally, as shown by \shortciteA{wang-etal-2023-element}, the part of their inferior performance could potentially be caused by the low-quality reference summaries used for evaluation. 

However, despite their current shortcomings, unsupervised methods show great potential for two reasons. First, the recent trends show summarization methods greatly benefit from large PLMs; even those not explicitly trained for summarization show acceptable performance in summarization \shortcite{lewis-etal-2020-bart}. As the performance of PLMs improves, the effectiveness of methods that use them also improves. Second, since collecting unlabeled data is more feasible than labeled data collection, and as shown by large PLMs \shortcite{radford2019language}, training on large amounts of data significantly improves the model's performance.
\subsection{Abstractive and Extractive}
From the recent trends and the number of research in abstractive and extractive methods, it is clear that during the past few years, the focus on summarization methods has shifted towards abstractive methods. This is due to the increasing number of labeled datasets, alongside the improvements in hardware capabilities, enabling training large models, and the introduction of efficient architectures and techniques such as copy mechanism \shortcite{vinyals-copy}. In unsupervised methods where labeled dataset is not often used, PLMs have been extensively used. Although some extractive methods use PLMs such as BERTSUM \shortcite{bertsum-DBLP:journals/corr/abs-1908-08345}, abstractive models use PLMs more often. 

However, even with the PLMs and other techniques, such as copy mechanism, used in summarization, both extractive and abstractive still struggle with their limitations. Extractive methods, being limited to input units, still lack fluency and do not perform well on datasets with abstractive reference summaries such as TIFU Reddit \shortcite{kim-etal-2019-abstractive}. Abstractive methods suffer from problems common in generation tasks such as hallucination \shortcite{rohrbach-etal-2018-object}, text degeneration \shortcite{text-degradation-DBLP:journals/corr/abs-1904-09751}, topic drift \shortcite{topic-drift-DBLP:journals/corr/abs-2009-06358} \shortcite{basu-roy-chowdhury-etal-2022-unsupervised}, and factual correctness \shortcite{factual1-10.5555/3504035.3504621} \shortcite{kryscinski-etal-2019-neural} \shortcite{maynez-etal-2020-faithfulness} \shortcite{oved-levy-2021-pass}. Additionally, long document summarization still challenges abstractive summarization \shortcite{liang-etal-2022-efficient}, and while the recent language models (e.g. GPT-4) have significantly increased context windows, their performance has yet to be studied and evaluated in summarization setting. While the more recent GPT-based models and other large-scale language models such as LLAMA 2 may have significantly improved in these areas, the main concern with these models, the high training and inference costs, still remains. Lastly, as shown by \shortciteA{suhara-etal-2020-opiniondigest}, extract-than-abstract approaches seem especially effective when dealing with long or multi-document summarization, as both abstractive and extractive techniques complement each other's weaknesses.

\subsection{Trends, Limitations and Future Directions}
Over the past few years, the research focus in unsupervised summarization has shifted multiple times. Before the transformer based models, extractive methods were often the state-of-the-art. The transformer technology shifted the focus towards abstractive models that usually employed encoder-decoder architectures. With the introduction of pretrained language models, most works focused on fine-tuning, modifying, or utilizing PLMs in the summarization pipeline. Over the past year, the surge of LLMs and their superior performance in many NLP tasks attracted research that focus on using LLMs for summarization. Recent LLMs not only generate high quality text, but also improve upon the previous problems of abstractive summarization such as hallucination and factual consistancy \footnote{https://github.com/vectara/hallucination-leaderboard}. However, there are challenges and limitations associated with LLMs in summarization and in general.  

\textbf{Training and Fine-tuning Costs} Unlike PLMs, LLMs are costly to train and fine-tune, and some of the best performing LLMs such as GPT-4 are not open-sourced. These factors make experimenting with LLMs (e.g. fine-tuning) impossible without access to powerful GPUs. This affects the type of research that many researchers might want to conduct due to imitated resources. Given that the current trend in summarization continues, there will be less focus on proposing novel architecture or training and fine-tuning in LLMs. Instead the research will focus around utilizing current state-of-the-art LLMs for NLP tasks. This is also evident in summarization papers in the past year where most of the works centered around prompt engineering and comparing the performance of various LLMs.

\textbf{Inference in Real-world Applications} Similarly, the computation resources required for LLMs also affects its real-world application. Currently, many of the LLM-powered application, such as chatbots, use LLMs hosted on servers. This inference costly for business and users. On the other hand, hosting current LLMs locally on consumer grade hardware is impossible due to their size. Therefore, approaches for reducing the size of current LLMs (such as quantization or sparsification) or training smaller task-specific language models through knowledge distillation using LLMs could prove useful.  

\textbf{Evaluation Metric} Another limitation of the summarization task is the commonly used evaluation metric for the task. Even though summarization techniques have greatly evolved and improved over the years, the metric used for evaluating summarization has stayed stagnant. In addition to its reproducibility and comparability problems \shortcite{grusky-2023-rogue}, ROUGE score is concerned with the n-grams overlap between candidates and summaries, and fails to consider the semantics of documents \shortcite{li-etal-2023-hear}. While language model evaluation metric is an active research area, they have not been commonly used by the community. 

\textbf{Dataset Quality and Domain} As shown by \shortciteA{wang-etal-2023-element}, the quality of reference summaries within datasets impacts the evaluation of summarization models and how they are benchmarked. The quality of summaries is affected by their factual consistency, hallucinations, and information redundancy. This makes evaluations done on low quality datasets unreliable. Additionally, a notable constraint lies in the scarcity of labeled summarization datasets beyond traditional news domains. Many crucial domains such as social media lack substantial, large-scale labeled datasets, hindering the development and evaluation of summarization techniques tailored to these specific contexts. This lack of diverse labeled data limits the possibility of evaluating and comparing summarization models across various domains.

\textbf{Long/Multi Document Summarization} Despite the potential of LLMs in processing long inputs, there is a lack of research focusing on experimenting with LLMs for multi-document and long-document summarization tasks. While much of the existing research concentrates on news articles and short textual inputs, the expanded context window of LLMs theoretically enables them to effectively summarize longer inputs. However, there is a notable gap in the literature when it comes to exploring and reporting the performance of LLMs on longer outputs. This lack of research limits our understanding of the capabilities of LLMs in handling longer documents. Thus, addressing this research gap is crucial for advancing the field of text summarization and leveraging the full potential of LLMs in handling different kinds of input. 


\section{Datasets and Evaluation Methods}
Most datasets focus on news summarization; the following are some of the most commonly used. CNN/Dailymail dataset \shortcite{cnn-dailymail/DBLP:journals/corr/HermannKGEKSB15}, Gigaword Corpus \shortcite{rush-etal-2015-neural}, DUC corpus \footnote{\url{https://duc.nist.gov/duc2004/}}, XSum \shortcite{narayan-etal-2018-dont}, Multi-News \shortcite{fabbri-etal-2019-multi}, and Newsroom \shortcite{grusky-etal-2018-newsroom}. For review summarization, Yelp reviews \shortcite{meansum-DBLP:journals/corr/abs-1810-05739}, Amazon reviews \shortcite{amazondataset-10.1145/2872427.2883037}, and SPACE hotel reviews \shortcite{angelidis-etal-2021-extractive} are commonly used. There are some datasets in the social media domain, mainly Twitter and Reddit; however, we noticed TIFU Reddit \shortcite{kim-etal-2019-abstractive} is the only one being actively used by researchers. Lastly, arXiv and PubMed are two datasets for scientific and medical papers \shortcite{cohan-etal-2018-discourse}.

Variants of the ROUGE \cite{lin-2004-rouge} F1 score (R1, R2, and RL) are standard evaluation metrics used in most of the existing work. Other automatic metrics, such as perplexity, are rarely used. Human evaluation is sometimes used as well, and while the criteria for human evaluation are not the same in every research, they mainly focus on coherence, coverage, fluency, informativeness, and redundancy. Despite its popularity, the ROUGE score has its limitations, such as favoring longer summaries \shortcite{schumann-etal-2020-discrete}, reproducibility and comparability \shortcite{grusky-2023-rogue}. Moreover, ROUGE only focuses on n-gram matches between the generated and reference summary that neither considers semantic similarity, nor considers other characteristics of an ideal summary such as fluency or factual consistency.  

Lastly, another line of research focuses on automatic evaluation metrics that aim to evaluate summaries in different aspects. BERTScore \shortciteA{bert-score}, BLEURT \shortciteA{sellam-etal-2020-bleurt}, and BARTScore \shortciteA{NEURIPS2021_e4d2b6e6} propose alternatives for ROUGE that use PLMs to score generated text with respect to its references.  
\shortciteA{wu-etal-2020-unsupervised} introduces an unsupervised approach based on BERT for evaluating summaries that does not require reference summaries. \shortciteA{kamal-eddine-etal-2022-frugalscore} a less resource intensive compared to PLM-based evaluation metrics that has comparable performance. Additionally with the prevalence of LLMs, works such as \shortciteA{zha-etal-2023-alignscore} focus on evaluating factual consistency. However one downside for these evaluation metrics is that they have yet to be adopted by other research, and despite its shortcomings, ROUGE still is the only commonly used evaluation metric for summarization task.  

\section*{Conclusion}
This survey reviews recent trends in unsupervised summarization. We explored the role of summarization and the importance of unsupervised methods. We covered various classifications in text summarization and offered our fine-grained taxonomy by examining the recent strategies and techniques. We further discussed the limitations of each approach, and compared their performance. Lastly, we mention popular datasets in each domain and touched upon evaluation metrics used in summarization.

\section*{Limitations}
In this survey, we focused on covering research from select venues over the past few years (check Paper Selection for details). We may have missed unsupervised methods from other venues or some research from the select venues. We also tried to cover critical previous works and baselines, but we could only cover the recent important research. We also did not run and test the models' performance ourselves and relied on metrics reported by the authors. However, we noticed that different research report varying ROUGE scores for baselines on the same dataset. As a result, comparisons among models may be inaccurate. 






\appendix


\vskip 0.2in
\bibliography{file}
\bibliographystyle{theapa}

\end{document}